\documentclass[article]{svjour3}

\usepackage{amsmath}
\usepackage{booktabs}
\usepackage{graphicx}
\usepackage{hyperref}
\usepackage{pdflscape}
\usepackage{multirow}
\usepackage{rotating}
\usepackage{spverbatim}
\usepackage{xcolor}

\newcommand{\ra}[1]{\renewcommand{\arraystretch}{#1}}

\newcommand{\figref}[1]{Figure~\ref{#1}}
\newcommand{\tabref}[1]{Table~\ref{#1}}

\title{Compressing Word Embeddings Using Syllables}
\author{Laurent Mertens \and Joost Vennekens}
\institute{KU Leuven, De Nayer Campus, Dept. of Computer Science\\
J.-P. De Nayerlaan 5\\
2860 Sint-Katelijne-Waver, Belgium\\
\and \\
Leuven.AI - KU Leuven Institute for AI\\
B-3000 Leuven\\
Belgium\\
\email{laurent.mertens@kuleuven.be}}
\date{\today}

\begin{document}
\maketitle

\abstract{This work examines the possibility of using syllable embeddings, instead of the often used $n$-gram embeddings, as subword embeddings. We investigate this for two languages: English and Dutch. To this end, we also translated two standard English word embedding evaluation datasets, WordSim353 and SemEval-2017, to Dutch. Furthermore, we provide the research community with data sets of syllabic decompositions for both languages. We compare our approach to full word and $n$-gram embeddings. Compared to full word embeddings, we obtain English models that are 20 to 30 times smaller while retaining 80\% of the performance. For Dutch, models are 15 times smaller for 70\% performance retention. Although less accurate than the $n$-gram baseline we used, our models can be trained in a matter of minutes, as opposed to hours for the $n$-gram approach. We identify a path toward upgrading performance in future work. All code is made publicly available, as well as our collected English and Dutch syllabic decompositions and Dutch evaluation set translations.}


\section{Introduction}
It is no exaggeration to state that there is a ``pre" and ``post" word embeddings era in the domain of Natural Language Processing (NLP). Despite their success, there are still a number of issues with these models. One such issue is their ever increasing size, which leads to substantial memory and processing power requirements.

A second issue is the problem of how to deal with words that are not present in the model, better know as the ``out of vocabulary" (OOV) problem. A typical way of dealing with this is by using models that do not consist of \emph{word} embeddings as such, but rather of subwords, most notably $n$-grams. A vector for a specific word is then created by combining the vectors for all $n$-grams present in this word.

The question still remains, however, whether $n$-grams are the best choice for the word constituents. In this work, we examine the possibility of approximating word embeddings by using syllabic decompositions, rather than $n$-grams. This might lead to greatly reduced model sizes (compared to the original ``full word" model), while retaining a lot of the expressivity of the original model. The intuition is that syllables are natural subwords, that by their very nature contain a lot of information. Similar to $n$-grams, they might serve as a means to overcome the OOV problem. Our work focuses on the English and Dutch languages, both Germanic languages, but with very different rules for making compound words. To enable our experiments, we address the lack of datasets for Dutch by translating two English word embedding evaluation sets, WordSim353 and SemEval-2017, to Dutch.

To summarize, we investigate two research questions: does the use of syllables isntead of $n$-grams provide a way to reduce model size with limited impact on embedding quality, and does the use of syllables instead of $n$-grams help to handle OOV words? 

First, we will explain how we created the Dutch translations of the English evaluation sets in \S\ref{s:dutch_eval_sets}. Following this, we discuss how we obtained datasets of syllabic decompositions in both languages, and used these to train syllable embeddings in \S\ref{s:syll_decomp_and_embeddings}. These embeddings are then evaluated against the aformentioned evaluation sets in \S\ref{s:no_splitter_results}. In \S\ref{s:syll_splitter}, we train a Transformer network to perform syllabic decomposition, and use this network to obtain embeddings for OOV words. We evaluate the performance of the Transformer network itself, as well as its impact on the embedding performance on the evaluation sets. Finally, we conclude in \S\ref{s:conclusion}.

The code, together with our syllabic decomposition datasets and Dutch evaluation sets can be found at \url{https://gitlab.com/EAVISE/lme/syllableembeddings}.

\section{Dutch embedding evaluation sets}
\label{s:dutch_eval_sets}
Reference evaluation sets for Dutch word embeddings are a rare thing. To help remedy this issue, we propose two Dutch translations of popular evaluation sets. One is the WordSim353 \cite{wordsim353} dataset, the other the SemEval-2017 English language \cite{semeval2017} dataset.

\subsection{WordSim353}
WordSim353 is a popular English language dataset used to evaluate word embeddings. It consists of 353 word pairs with corresponding human-assigned similarity judgements in the form of scores on a scale of 0 to 10. To convert the dataset to Dutch, the first author simply translated all pairs, taking into account the similarity score in case several translations were possible. The translated dataset was not rescored; the original scores were kept.

\subsection{SemEval-2017}
SemEval is shorthand for the International Workshop on Semantic Evaluation \cite{semeval_url}, a series of workshops focusing on the evaluation of NLP tasks. In the 2017 edition, one of these tasks was ``Multi­lingual and Cross­-lingual Semantic Word Similarity". When we refer to ``SemEval-2017", we refer to this particular task and its associated datasets. For the purpose of this task, an English language dataset was created consisting of 500 word pairs with corresponding human-assigned similarity judgements using a five-point Likert scale. This dataset was then translated to several languages following a standard procedure to allow multilingual and cross-lingual evaluation.

To create a Dutch language version of the SemEval-2017 dataset, we followed this same procedure, as described in \S 2.1.2 of the original paper, with some small deviations. Summarized, the performed steps are:
\begin{enumerate}
\item The English words were translated to Dutch, but only by one person (as opposed to two in the original paper).
\item All word pairs were then annotated by three different annotators using the five-point Likert scale defined in Table 2 of the original paper. Contrary to the paper, we did not allow fractional values in between the defined points in the scale. 
\item For each annotator, we looked at which scores deviated more than 1 point from the average of the other two annotators. Each annotator was then asked to reassess these particular word pairs and possibly update their score.
\item After this rescoring step, the final similarity score for each word pair was calculated as the arithmetic mean of the final annotations.
\end{enumerate}

The Pearson correlation scores between annotators is shown in \tabref{tb:nl_semeval_ann_corr}. Our final correlation value of 0.836 deviates from the values close to 0.9 for all languages found in \cite{semeval2017}, but this can be explained by the fact that we did not allow inter-value annotations. The Pearson correlation between the original English and our Dutch sets of 0.873 is indicative of a very good overall match between both sets.

\begin{table}
\centering
\caption{Pairwise Pearson correlation between annotators. ``Ann.P." = Annotator Pair, ``Avg" = Average.}
\label{tb:nl_semeval_ann_corr}
\ra{1.1}
\begin{tabular}{ccccc}
& Ann.P. 1 & Ann.P. 2 & Ann.P. 3 & Avg. \\
\midrule
Initial scores & 0.828 & 0.784 & 0.781 & 0.798 \\
Revised scores & 0.864 & 0.846 & 0.796 & 0.836
\end{tabular}
\end{table}

\subsection{Baseline embedding scores}
\label{ss:baseline_scores}
As a baseline for all of our experiments, we use the ConceptNet Numberbatch v19.08 word embeddings \cite{conceptnet}. To check performance against the $n$-gram approach, we also include results obtained by using $n$-gram models trained using the approach of \cite{sasaki2019}.

Numberbatch (NB) contains embeddings in numerous languages, including English and Dutch. More specifically, there are embeddings for 516,783 English words and 190,221 Dutch words, which achieve a Spearman correlation of 0.815 on the English WordSim353 dataset, 0.683 on our Dutch translation, and 0.645 and 0.662 on the English and Dutch SemEval 2017 datasets respectively. The number of unique words in each evaluation set is depicted in \tabref{tb:nb_words_per_evalsets}.

\begin{table}
\centering
\caption{Number of unique words per evaluation set and language.}
\label{tb:nb_words_per_evalsets}
\ra{1.1}
\begin{tabular}{ccc}
Language & WordSim353 & SemEval 2017 \\
\midrule
English & 437 & 915 \\
Dutch & 424 & 910
\end{tabular}
\end{table}

For the $n$-gram baseline, we separately trained English and Dutch language models using the source code provided by the authors of \cite{sasaki2019}, and using the Numberbatch embeddings as source embeddings. Note that we also generated new $n$-gram min/max dictionaries for each language. We then used these models to generate embeddings for all unique words present in the each evaluation set. We trained two models for each language with different number of parameters. The first model uses 100,000 $n$-grams, resulting in 30,000,000 parameters, while the second used only 15,000 $n$-grams, or 4,500,000 parameters, which is more in line with the size of our models. We refer to these models as ``S.2019.a" and ``S.2019.b" respectively.

\section{Syllabic decompositions and embeddings}
\label{s:syll_decomp_and_embeddings}
We gathered a list of syllabic decompositions in Dutch and English, by:
\begin{itemize}
\item Parsing the Dutch Wiktionary. We used the backup dump of the 1st of November, 2019. These dumps are available for download from \cite{wiktdump}, although older dumps are removed in favor of more recent ones. Many entries in this Wiktionary contain the syllabic decomposition of the word the page relates to. Only those pages with titles that, when lowercased, contained only the characters shown in \tabref{tb:filter_chars} were retained.
\item Scraping two public websites for English, ``How Many Syllables" \cite{howmanysyllables} and ``Syllable Count" \cite{syllablecount}, since the English wiktionary contains no syllabic decompositions. For scraping, we followed the ordering of the most frequent words in the ConceptNet embeddings. As such, we did not filter the words as we did for the Dutch dataset.
\end{itemize}
\begin{table}
\center
\caption{List of allowed characters when filtering words. Whitespace is only used for readability, but is not an allowed character itself.}
\label{tb:filter_chars}
\ra{1.1}
\begin{tabular}{c}
\toprule
abcdefghijklm \\
nopqrstuvwxyz \\
0123456789 \\ 
\"a\`a\'a \c{c} \"e\`e\'e\^e \\
\"i\'i\^i \~n \"o\'o \"u\'u - \\
\bottomrule
\end{tabular}
\end{table}

\tabref{tb:syll_scrape_stats} contains the details on how many words were collected, as well as how many unique syllables these contain. There are a lot more Dutch tokens than English ones, even after data clean up, partly because these contain plurals which the English set does not. However, this does not suffice however to fully explain the discrepancy in size. Of course, to increase the size of the English dataset one could scrape more, but we noticed that at this point, additional scraping only adds a fraction of the words contained in ConceptNet. In other words, it seems that most words that are contained in ConceptNet and that are not yet scraped are simply not present on either of the two scraped websites.

\begin{table}
\centering
\caption{Syllable statistics per language.}
\label{tb:syll_scrape_stats}
\ra{1.1}
\begin{tabular}{ccc}
Language & \# Tokens & \# Uq. syllables \\
\midrule
English & 71,387 & 17,444 \\
Dutch & 357,784 & 18,929
\end{tabular}
\end{table}

We used the words for which we have a syllabic decomposition as training set to extract the embeddings from the original ConceptNet embeddings. We do this in two ways: using a vanilla approach, and using a slightly modified approach that uses attention.

Note however that not all words for which we have a syllabic decomposition also appear in the ConceptNet embeddings. In other words: there are many (unique) syllables that only appear in words for which we do not have an embedding, and hence, there is no point in training an embedding for those syllables. We only train embeddings for those syllables contained in words that overlap between both datasets. For English, this amounts to 66,674 words constituting 16,032 unique syllables, while for Dutch, 90,017 words overlap for a total of 9,814 unique syllables.

\subsection{Vanilla approach}
For the vanilla approach, we define the syllabic embedding $E(x)$ of word $x$ as: 
\begin{equation}
E(x) = \frac{\sum_{s_i \in S(x)} s_i}{\left\lVert \sum_{s_i \in S(x)} s_i \right\rVert _2},
\end{equation}
where $S(x)$ represents the syllables of which word $x$ consists, and $s_i$ is the embedding of syllable $i$. Training is implemented in PyTorch by storing the embeddings in a \texttt{torch.nn.Parameter} tensor and a forward pass that sums the appropriate syllable embeddings by multiplying an $n$-hot encoded vector, with 1's at those indices corresponding to the appropriate syllables, with this tensor. Optimization is done using an MSE loss. 

\subsection{Attention approach}
\begin{figure}[!htbp]
\centering
\includegraphics[width=10cm]{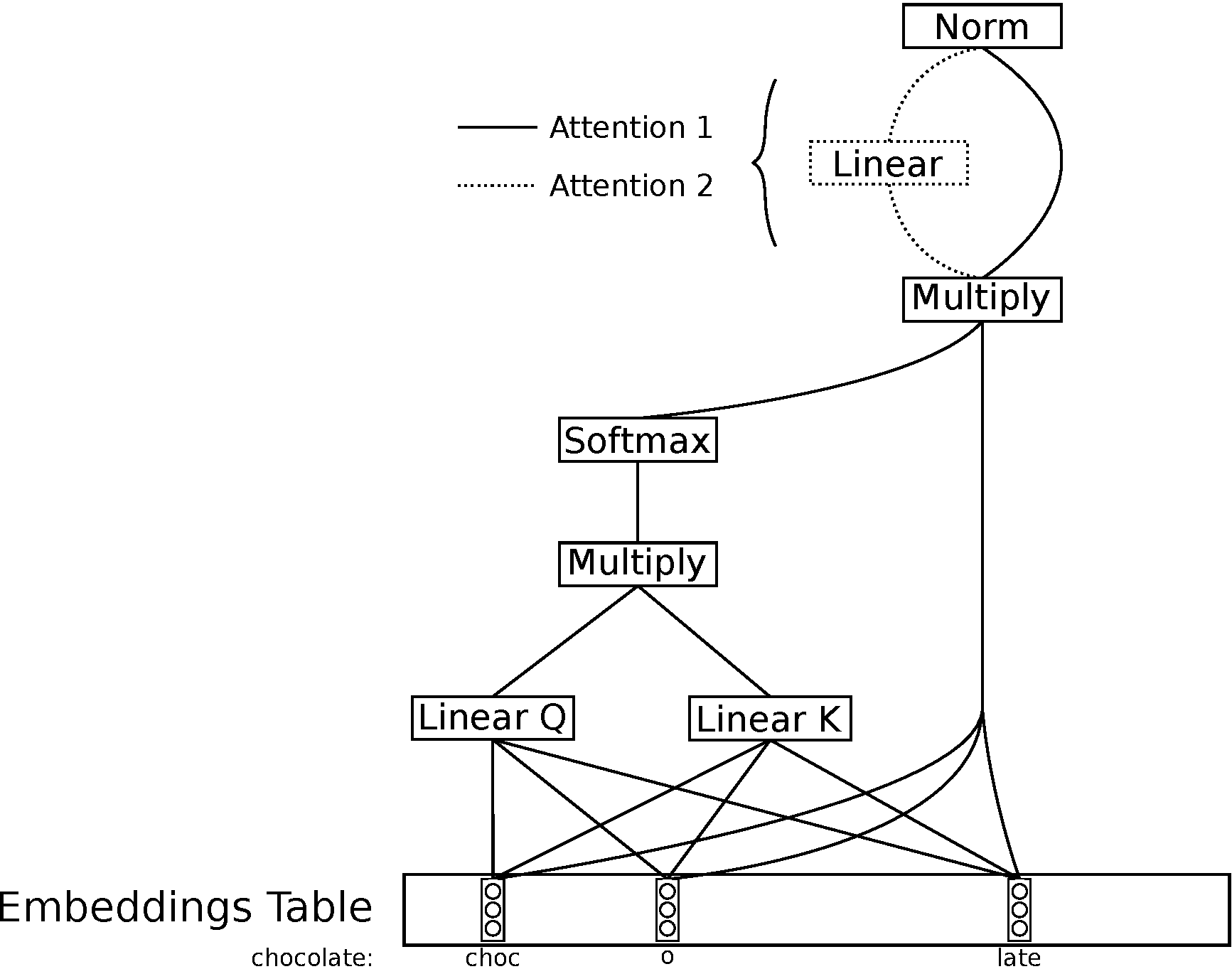}
\caption{Graphical illustration of the topology of networks Attention 1 and 2.}
\label{fig:network_topology}
\end{figure}

At first, we experimented with using the encoder part of a Transformer \cite{transformer} for learning the syllable embeddings, but we found that adding positional encoding had a negative effect, despite repeated attempts at getting the balance in weighting right (so as not to let the positional information dominate the values). As such, we decided, very much like \cite{sasaki2019}, to only use an attention mechanism. In the remainder, we refer to this network as ``Attention 1". We also used a variant that uses embeddings of lower dimension ($D=200$) than the original embeddings, which are of $D=300$, and that uses an extra linear layer to expand these embeddings back to a $300$-dimensional vector. This model will be referred to as ``Attention 2". Both of these topologies are shown in \figref{fig:network_topology}. All models were trained using an MSE loss, Adam optimizer using the AMSGrad variant, $\beta_1 = 0.90$, $\beta_2 = 0.999$, $\epsilon = 1^{-10}$ and an original learning rate $lr_0 = 0.05$ with update function
\begin{equation}
lr(e) = \frac{lr_0}{e/2 + 1},
\end{equation}
with $e$ the epoch.

\subsection{Start/end tokens}
\label{ss:start_end_tokens}
A common trick used to increase performance of subword models is to add explicit start- and end-token variants of tokens. What this means is that when a subword appears at the beginning of a word, it is typically prepended by a special character to encode this, and analogous for when a subword appears at the end of a word. For our case, as an example, consider the word ``apple". Its syllabic decomposition is ``ap-ple". Adding start- and end-tokens in this case would change this to ``\$ap-ple\#", where we prepend a syllable by `\$' to indicate it's at the start of the word, and analogous for `\#'. One-syllable words like, e.g., ``box" then become ``\$box\#". This allows the model to train up to four different variants for a same syllable (e.g., for ``box" we get ``box", ``\$box", ``box\#" and ``\$box\#") representing different positions of the syllable in a word.

For the Attention 1 and 2 models, we also experimented with only adding start/end variants for a fraction of all syllables. For this, we ordered syllables in descending order of how frequently they appeared at the start or end of a word, and would only add variants for the top $x$\% for both cases--start and end--separately, with $x$ a parameter.

There is no point in adding variants for syllables that do not appear at the start or end of a word, since no embeddings can be trained for these variants.

Here, the choice to investigate English and Dutch leads to an interesting observation. Both languages share West Germanic as common root, but differ in at least one major way: in Dutch words are compounded, while not so in English. Consider, e.g., the English phrase ``car insurance company", which in Dutch, using the same three basic words, is compounded into ``autoverzekeringsmaatschappij". The effect is further exacerbated by the use of diminutives: ``little car insurance company" is ``autoverzekeringsmaatschappijtje". Because of this, we expect the potential benefit of using start/end tokens to be less for Dutch than for English.

\subsection{Analyzing syllable sparseness}
\label{ss:syllable_sparseness}
To get a better understanding of the distribution of syllables over the vocabulary, we performed two statistical analyses.

\begin{figure}[!htbp]
\centering
\includegraphics[width=11cm]{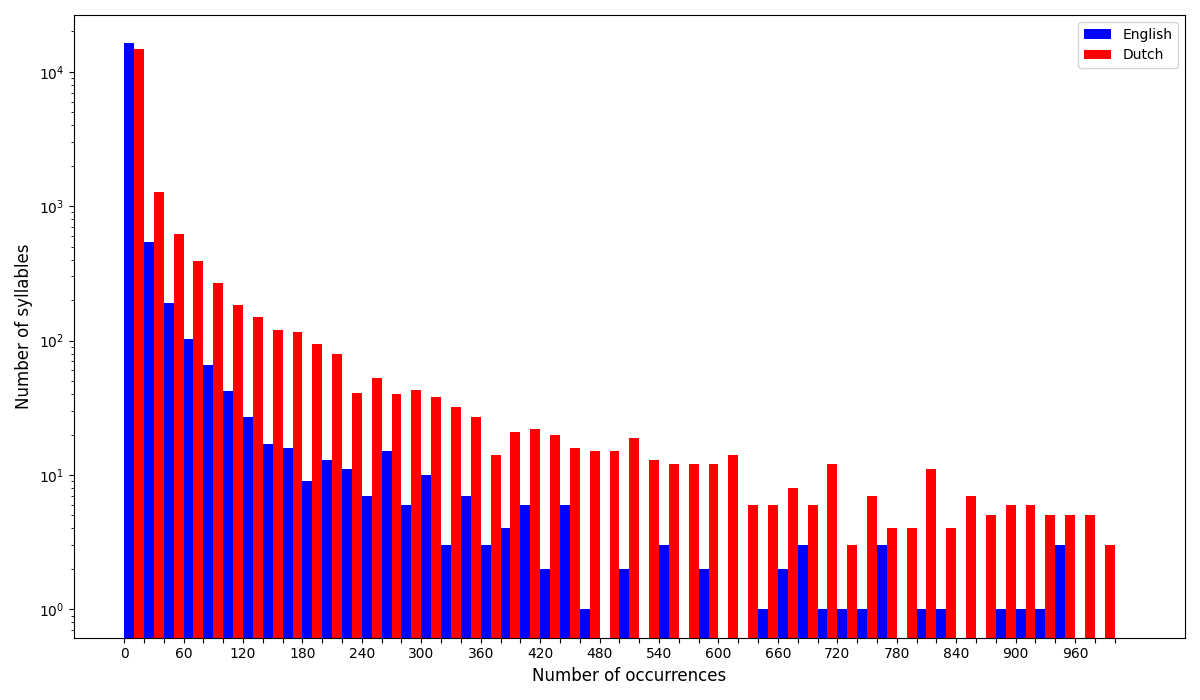}
\caption{Number of syllables appearing $x$ times. Bin width = 20.}
\label{fig:syll_bin_counts}
\end{figure}

First, we created a histogram of the number of times syllables appear in the vocabulary. This is shown in \figref{fig:syll_bin_counts}. The chart is limited to syllables that appear $<=1000$ times. There were 20 (0.11\%) and 186 (0.98\%) syllables that exceeded that limit for English and Dutch respectively. The graph shows a very clear concentration of syllables within the first bin, i.e., with a number of occurrences $\in [1, 20]$. For English, this first bin contains no less than 93.4\% of all syllables, while for Dutch the number is still an important 78.4\%. In other words, the vast majority of syllables are rather rare.

\begin{figure}[!htbp]
\centering
\includegraphics[width=9cm]{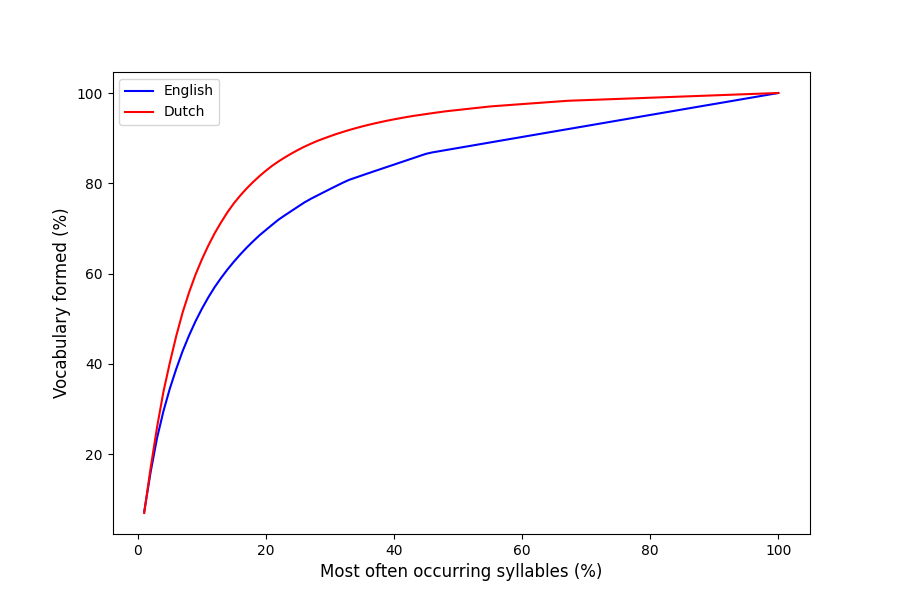}
\caption{Percentage of words that can be formed with top $x$\% of syllables.}
\label{fig:syll_top_x}
\end{figure}

We also looked at how many words contain only syllables from the top $x$\% most frequent syllables. The result is shown in \figref{fig:syll_top_x}. This graph shows that the vast majority of words consist of only a limited set of syllables. For English, the top 40\% of syllables accounts for almost 80\% of words, while for Dutch that number is further reduced to the top 20\%.

This analysis shows that syllable sparseness is potentially an issue, in that it will be very difficult to learn good embeddings (i.e., that generalize well) for syllables that have only very few occurrences. At the same time, however, the syllables for which there are many occurences account for the vast majority of words, so the impact of the data sparseness on the average word-level accuracy may be limited.

\subsection{Testing OOV capability}
%
%
One of our goals is to investigate whether syllable embeddings can be used to interpret OOV words, i.e., words that do not appear in the training set. Of course, this can only be succesful for OOV words that contain syllables that also appear in words that \emph{are} part of the training set. Because our analysis of Section \ref{ss:syllable_sparseness} shows that many syllables appear only in very few words, we have to take this into account when generating a suitable training set. Therefore, we decided to remove all words from the training set that entirely consist of syllables that occur at least $x$ times in the entire dataset, with $x$ a parameter. The tables in this document contain results for models with $x=3$ and $x=10$ in columns using evaluation set names ending with ``/3ct" and ``/10ct" respectively. \tabref{tb:oov_words_in_eval} contains the number of OOV words in the evaluation sets (i.e., the number of removed words) for each setup present in the evaluation tables. Recall that the number of words per evaluation set can be found in \tabref{tb:nb_words_per_evalsets}.

To compare against the $n$-gram approach, we also trained a ``S.2019.b" model (see \ref{ss:baseline_scores}) by removing all words present in the evaluation sets from the training corpus. In other words, for these models, all evaluation words become OOV words.

\begin{table}[!htbp]
\centering
\caption{Number of OOV words per evaluation setup; `SE' = start/end tokens.}
\label{tb:oov_words_in_eval}
\begin{tabular}{cccccccccc}
\multirow{3}{*}{SE} & \multicolumn{4}{c}{WordSim353} && \multicolumn{4}{c}{SemEval 2017} \\
& \multicolumn{2}{c}{EN} & \multicolumn{2}{c}{NL} && \multicolumn{2}{c}{EN} & \multicolumn{2}{c}{NL} \\
& 3ct & 10ct & 3ct & 10ct && 3ct & 10ct & 3ct & 10ct \\
\midrule
0\%   & 410 & 320 & 399 & 362 && 674 & 513 & 681 & 614 \\
10\%  & 372 & 238 & 364 & 327 && 609 & 362 & 641 & 569 \\
20\%  & 314 & 208 & 330 & 256 && 517 & 319 & 587 & 439 \\
50\%  & 284 & 201 & 298 & 240 && 475 & 311 & 524 & 389 \\
75\%  & 282 & 200 & 287 & 238 && 471 & 309 & 498 & 380 \\
100\% & 282 & 200 & 285 & 238 && 470 & 307 & 492 & 377 
\end{tabular}
\end{table}



\subsection{A note on multiword expressions}
In case of multiword expressions, our system computes a (normalized) embedding for each separate word, which are then combined by adding the individual embeddings and normalizing the sum.

\section{Results}
\label{s:no_splitter_results}
Tables~\ref{tb:en_emb_results} and \ref{tb:nl_emb_results} show the results when evaluating word embeddings trained with our models on the WordEval353 and SemEval 2017 datasets for English and Dutch respectively. Note that we ignore casing, and that at inference time, we look up the syllabic decomposition of a word in our database. Hence, we are only concerned about the performance of combining syllabic embeddings into full word embeddings, and at this stage do not consider the problem of how to obtain the syllabic decomposition for words for which the decomposition is not known in our database.

What is immediately apparent from this is that the attention approach is vastly superior to the vanilla approach at barely any additional expense in terms of parameters. Because of this, we did not do additional experiments with ``/3ct" and ``/10ct" models or fractional start/end tokens for the vanilla model, but only for the Attention models.

Turning to the Attention models, it is clear that the impact of the attention mechanism on performance is significant, yet so is also the reduction in model size compared to the NumberBatch baseline. Results for English are significantly better than for Dutch, with the best Attention 1 model result for WordSim353 at 77.5\% of the baseline for a model compression ratio of 23.1, and the SemEval 2017 result at 83.4\% of the baseline for a model compression ratio of 19.3. A full list of performance and compression ratio results for the best performing model in each class for both languages can be found in \tabref{tb:performance-to-compression}.

Furthermore, the difference in performance between the best $D=300$ (Attention 1) and $D=200$ (Attention 2) model is not always that big, despite a 33\% reduction in model size. For the Dutch language, we even have the surprising result of the lower dimensional model performing the best.

\begin{landscape}
\begin{table}[!htbp]
\centering
\caption{Accuracy results for different embedding methods for English. Model names in \emph{emphasis}, `NB' = baseline, `S.2019.a/b' are the Sasaki et al. embeddings, `SE' = start/end tokens, `MW' = missing words, the number of words for which no embedding is present or could be formed. Best results per model in \bf{bold}.}
\label{tb:en_emb_results}
\ra{1.2}
\begin{tabular}{@{}rccccccccccc@{}}
& \# Params. && MW & WS353 & WS353/3ct & WS353/10ct && MW & SE2017 & SE2017/3ct & SE2017/10ct \\
\midrule
\emph{Baseline} &&&&&&&&&&&  \\
NB & 155,034,600 && 0 & 0.815 & -- & -- && 134 & 0.645 & -- & -- \\
S.2019.a & 30,000,000 && 3 & 0.793 & -- & -- && 20 & 0.669 & -- & -- \\
S.2019.b & 4,500,000 && 3 & 0.697 & 0.679$^*$ & -- && 20 & 0.564 & 0.515$^*$ & -- \\
\midrule
\emph{Vanilla} &&&&&&&&&&& \\
SE: No & 4,809,600 && 2 & 0.416 & -- & -- && 75 & 0.401 & -- & -- \\
SE: 100\% & 7,833,000 && 2 & \bf{0.503} & -- & -- && 75 & \bf{0.505} & -- & -- \\
\midrule
\emph{Attention 1} &&&&&&&&&&& \\
SE: No & 4,989,900 && 2 & 0.513 & 0.343 & 0.406 && 75 & 0.437 & 0.343 & 0.354 \\
SE: 10\% & 5,842,800 && 2 & 0.604 & 0.367 & 0.466 && 75 & 0.500 & 0.385 & 0.438 \\
SE: 25\% & 6,697,500 && 2 & \bf{0.632} & 0.402 & 0.480 && 75 & 0.518 & 0.410 & 0.457 \\
SE: 50\% & 7,426,800 && 2 & 0.624 & 0.400 & 0.472 && 75 & 0.516 & \bf{0.446} & \bf{0.461} \\
SE: 75\% & 7,796,700 && 2 & 0.631 & 0.399 & 0.489 && 75 & 0.528 & 0.427 & 0.457 \\
SE: 100\% & 8,013,300 && 2 & 0.629 & \bf{0.404} & \bf{0.501} && 75 & \bf{0.538} & 0.437 & 0.457 \\
\midrule
SE: No & 3,346,900 && 2 & 0.498 & 0.342 & 0.401 && 75 & 0.456 & 0.331 & 0.340 \\
SE: 10\% & 3,915,500 && 2 & 0.580 & 0.363 & 0.431 && 75 & 0.488 & 0.394 & 0.409 \\
SE: 25\% & 4,485,300 && 2 & \bf{0.600} & 0.358 & 0.443 && 75 & 0.507 & 0.375 & \bf{0.448} \\
SE: 50\% & 4,971,500 && 2 & 0.564 & 0.377 & \bf{0.449} && 75 & \bf{0.519} & 0.411 & 0.440 \\
SE: 75\% & 5,218,100 && 2 & 0.588 & \bf{0.388} & 0.412 && 75 & 0.517 & 0.419 & 0.438 \\
SE: 100\% & 5,362,500 && 2 & 0.598 & 0.380 & 0.447 && 75 & 0.503 & \bf{0.420} & 0.425 \\
\bottomrule
\multicolumn{12}{l}{$^*$\scriptsize{These models do not follow the ``3ct" rule, but were trained by removing \emph{all} evaluation words from the training set. }}
\end{tabular}
\end{table}
\end{landscape}

\begin{landscape}
\begin{table}[!htbp]
\centering
\caption{Accuracy results for different embedding methods for Dutch. Model names in \emph{emphasis}, `NB' = baseline, `S.2019.a/b' are the Sasaki et al. embeddings, `SE' = start/end tokens, `MW' = missing words, the number of words for which no embedding is present or could be formed. Best results per model in \bf{bold}.}
\label{tb:nl_emb_results}
\ra{1.2}
\begin{tabular}{@{}rccccccccccc@{}}
& \# Params. && MW & WS353 & WS353/3ct & WS353/10ct && MW & SE2017 & SE2017/3ct & SE2017/10ct \\
\midrule
\emph{Baseline} &&&&&&&&&&& \\
NB & 57,066,300 && 13 & 0.683 & -- & -- && 146 & 0.662 & -- & -- \\
S.2019.a & 30,000,000 && 3 & 0.697 & -- & -- && 20 & 0.635 & -- & -- \\
S.2019.b & 4,500,000 && 3 & 0.589 & 0.516$^*$ & -- && 20 & 0.542 & 0.462$^*$ & -- \\
\midrule
\emph{Vanilla} &&&&&&&&&&& \\
SE: No & 2,944,200 && 20 & 0.334 & -- & -- && 179 & 0.377 & -- & -- \\
SE: 100\% & 5,720,100 && 20 & \bf{0.385} & -- & -- && 179 &\bf{0.399} & -- & -- \\
\midrule
\emph{Attention 1} &&&&&&&&&&& \\
SE: No & 3,124,500 && 20 & 0.428 & 0.375 & 0.372 && 174 & 0.393 & \bf{0.369} & 0.372 \\
SE: 10\% & 3,596,400 && 20 & 0.453 & 0.396 & 0.394 && 174 & 0.415 & 0.351 & 0.370 \\
SE: 25\% & 4,289,700 && 20 & 0.474 & \bf{0.412} & \bf{0.412} && 174 & 0.417 & 0.346 & 0.380 \\
SE: 50\% & 5,133,600 && 20 & 0.482 & 0.386 & 0.386 && 174 & \bf{0.448} & 0.368 & 0.391 \\
SE: 75\% & 5,620,200 && 20 & \bf{0.484} & 0.349 & 0.404 && 174 & 0.443 & 0.335 & 0.400 \\
SE: 100\% & 5,900,400 && 20 & 0.470 & 0.386 & 0.391 && 174 & 0.424 & 0.342 & \bf{0.402} \\
\midrule
\emph{Attention 2} &&&&&&&&&&& \\
SE: No & 2,103,300 && 20 & 0.428 & 0.395 & 0.377 && 174 & 0.418 & \bf{0.386} & 0.378 \\
SE: 10\% & 2,417,900 && 20 & 0.432 & 0.363 & 0.346 && 174 & 0.418 & 0.367 & 0.360 \\
SE: 25\% & 2,880,100 && 20 & 0.473 & \bf{0.384} & \bf{0.390} && 174 & 0.438 & 0.362 & 0.400 \\
SE: 50\% & 3,442,700 && 20 & 0.462 & 0.358 & 0.385 && 174 & 0.458 & 0.360 & 0.397 \\
SE: 75\% & 3,767,100 && 20 & \bf{0.485} & 0.373 & 0.382 && 174 & 0.452 & 0.351 & \bf{0.405} \\
SE: 100\% & 3,953,900 && 20 & 0.472 & 0.368 & 0.375 && 174 & \bf{0.462} & 0.369 & 0.393 \\
\bottomrule
\multicolumn{12}{l}{$^*$\scriptsize{These models do not follow the ``3ct" rule, but were trained by removing \emph{all} evaluation words from the training set. }}
\end{tabular}
\end{table}
\end{landscape}

\begin{table}[!htbp]
\centering
\caption{Performance and compression ratio as compared to baseline for best performing models per type for English and Dutch.}
\label{tb:performance-to-compression}
\ra{1.2}
\begin{tabular}{@{}l c cc c cc@{}}
&& \multicolumn{2}{c}{English} && \multicolumn{2}{c}{Dutch} \\
\cline{3-4} \cline{6-7}
&& WS353 & SE2017 && WS353 & SE2017 \\
\midrule
\multicolumn{7}{l}{\emph{Performance (\%)}} \\ 
Vanilla     && 61.7 & 78.2 && 56.4 & 60.3 \\
Attention 1 && 77.5 & 83.4 && 69.3 & 67.2 \\
Attention 2 && 73.6 & 80.5 && 69.3 & 69.5 \\
\midrule
\multicolumn{7}{l}{\emph{Compression (factor)}} \\ 
Vanilla     && 19.8 & 19.8 && 10.0 & 10.0 \\
Attention 1 && 23.1 & 19.3 && 10.2 & 11.1 \\
Attention 2 && 34.6 & 31.2 && 15.1 & 14.4 \\
\end{tabular}
\end{table}

Also interesting to note is that more often than not the 100\% start/end token model is outperformed by a lower percentage model. We hypothesize that this is due to the increased sparseness when adding start/end variants. Many syllables already have a low incidence count, and splitting each syllable into multiple variants reduces this number ever further. This causes our smaller models to achieve better or equal performance as the largest models.

Our intuition that the start/end tokens would be less informative for Dutch compared to English, as explained in \S\ref{ss:start_end_tokens}, is also confirmed by these results, as evidenced by a larger gap between the 0\% and 100\% results for the Attention models for English than for Dutch. 
With regards to our first research question, we can conclude that syllables indeed offer a useful way of reducing model size. However, they are outperformed by the $n$-grams: the S.2019.b model typically has a 5--10\% higher accuracy for a comparable number of parameters. 
Both S.2019 models outperform the syllable-based models for both languages, and moreover are able to resolve more words (i.e., there are less ``missing words" as evidenced by the ``MW" column, except for performance on the English WordSim353 evaluation set). It is interesting to note though that training a syllable model takes a few minutes, while training of the S.2019 models ranged from 4 hours (S.2019.b, Dutch) to 16 hours (S.2019.a, English).

Performance clearly takes a big hit when turning to the OOV problem, as evidenced by the performance of the ``/3ct" and ``/10ct" models. The most probable explanation for this is two-fold:
\begin{itemize}
\item Data sparseness leading to embeddings that do not generalize well.
\item The embedding for a word being constructed from one single decomposition.
\end{itemize}
Typically, when using subword embeddings, an embedding for a word is constructed by taking into account all the possible ways the word can be formed from the subwords. However, in our case, this is made practically infeasible by the quickly rising number of possible decompositions as words get longer. E.g., the word ``aandeelhoudersvergadering" has 322,560 possible decompositions using our syllable set, and ``aansprakelijkheidsverzekeringen" has no less than 3,179,520 possible decompositions. 
We therefore only consider the real decomposition (as determined in the way discussed in \S\ref{s:syll_decomp_and_embeddings}) of each word. When comparing OOV performance between the syllable and $n$-gram approaches, we see that the S.2019.b model takes the biggest accuracy hit for Dutch, while for the syllable approach it is the other way around. Especially noteworthy is that, although the S.2019.b model is clearly superior for English (i.e., less accuracy loss compared to syllable approach), the performance hit taken by both approaches for Dutch is very similar, typically in the order of 8\%.

For completeness sake it should be pointed out that comparing our model sizes to the ConceptNet embeddings isn't entirely accurate, as the ConceptNet embeddings contain embeddings for words containing syllables for which we cannot train an embedding. In other words: our models would not allow to reproduce the entire vocabulary of the ConceptNet embeddings. A fairer comparison would be to only look at that subset of words from ConceptNet for which we can compute an embedding, but this is a Catch-22 situation, as it would require us to known the syllabic decomposition of all words.

\section{Syllabic decomposition by Transformer}
\label{s:syll_splitter}
The problem with using syllables as the constituent parts of a word is that one needs to know the syllabic decomposition of the word in order to be able to create an embedding for it. One could store a mapping into memory, as we did for the results discussed in Section \ref{s:no_splitter_results}, but then the problem remains for words that are not contained in the mapping. Alternatively, it should be possible to just implement the existing rules for syllabic decomposition, were it not that one will rapidly find oneself mirred into a web of exceptions and rules that are not easy to implement straightforwardly. In order to solve this problem, we trained several neural networks to split words into syllables.

To create a training set for Dutch, 250,000 words were randomly chosen out of the 357,784 words that comprise the Wiktionary dataset. These were then filtered using the characters displayed in \tabref{tb:filter_chars}, after which 227,557 tokens remained as training data. The reason this filtering eliminates words, although we used the same characters to filter as earlier, is that originally in \S\ref{s:syll_decomp_and_embeddings} we did not take into account the syllables, but only looked at the lowercased word (or Wikipedia page title to be precise). This time, we also looked at the syllables themselves, meaning that words with syllables containing uppercase letters were removed. Mainly, this allows to eliminate named entities and abbreviations. To make this filtering scheme more concrete, consider the following example. The Dutch wiktionary contains a page titled ``pick-up" describing this particular word. For the filtering scheme used in \ref{s:syll_decomp_and_embeddings}, we checked this (lowercased) word against the characters in \tabref{tb:filter_chars}, and would decide not to parse this page because of the presence of the `-' character. For the current purposes, we would not look at the page title, but at the syllabic decomposition itself (displayed on the page), which in this case happens to be identical, and reject the word on the ground that its syllabic decompostion contains forbidden characters.

All in all, the Dutch evaluation set consists of the 107,784 remaining words, which the filtering reduces to 98,190 words. Analogously, for English 55,000 words were randomly chosen, which were filtered down to 50,866, leaving 15,084 words for evaluation after filtering the remaning words.

\begin{landscape}
\begin{table}[!htbp]
\caption{Accuracy (\%) of Dutch and English syllablic decomposition Transformer networks. `embedding' and `hidden' both refer to dimensions. Best results in bold.}
\label{tb:syll_split_results}
\centering
\ra{1.2}
\begin{tabular}{@{}rrrrcrrrcrrr@{}}
\toprule English & \multicolumn{3}{c}{$embedding = 16$} & \phantom{abc}& \multicolumn{3}{c}{$embedding = 32$} & \phantom{abc} & \multicolumn{3}{c}{$embedding = 64$}\\
\cmidrule{2-4} \cmidrule{6-8} \cmidrule{10-12}& $heads=4$ & $heads=8$ & $heads=16$ && $heads=4$ & $heads=8$ & $heads=16$ && $heads=4$ & $heads=8$ & $heads=16$ \\
\midrule$layers=1$\\
$hidden=64$ & 71.0 & -- & -- && 74.3 & 75.2 & -- && 76.5 & 77.0 & 77.2 \\
$hidden=128$ & 70.5 & -- & -- && 74.6 & 75.0 & -- && 75.6 & 76.5 & 77.4 \\
$hidden=256$ & 71.1 & -- & -- && 75.1 & 75.1 & -- && 75.7 & 77.0 & 78.1 \\
$layers=2$\\
$hidden=64$ & 73.7 & -- & -- && 76.9 & 76.7 & -- && 76.5 & 77.6 & 78.1 \\
$hidden=128$ & 74.5 & -- & -- && 76.7 & 77.6 & -- && 77.5 & 78.1 & 77.9 \\
$hidden=256$ & 74.1 & -- & -- && 77.0 & 77.2 & -- && 78.2 & \textbf{78.6} & 78.4 \\
\bottomrule
\toprule Dutch & \multicolumn{3}{c}{$embedding = 16$} & \phantom{abc}& \multicolumn{3}{c}{$embedding = 32$} & \phantom{abc} & \multicolumn{3}{c}{$embedding = 64$}\\
\cmidrule{2-4} \cmidrule{6-8} \cmidrule{10-12}& $heads=4$ & $heads=8$ & $heads=16$ && $heads=4$ & $heads=8$ & $heads=16$ && $heads=4$ & $heads=8$ & $heads=16$ \\
\midrule$layers=1$\\
$hidden=64$ & 85.7 & -- & -- && 87.8 & 88.6 & -- && 89.4 & 89.5 & 89.9 \\
$hidden=128$ &  85.7 & -- & -- && 88.8 & 89.0 & -- && 89.9 & 90.2 & 90.0 \\
$hidden=256$ & 86.1 & -- & -- && 88.2 & 88.6 & -- && 90.1 & \textbf{90.3} & 90.1 \\
$layers=2$\\
$hidden=64$ & 86.8 & -- & -- && 88.4 & 87.7 & -- && 89.0 & 90.0 & 89.3 \\
$hidden=128$ & 87.1 & -- & -- && 89.3 & 88.8 & -- && 89.6 & 89.4 & 89.3 \\
$hidden=256$ & 86.7 & -- & -- && 88.7 & 88.9 & -- && 89.6 & 90.0 & 90.0 \\
\bottomrule
\end{tabular}
\end{table}
\end{landscape}

\begin{landscape}
\begin{table}[!htbp]
\centering
\caption{Accuracy results for different embedding methods: English. Model names in \emph{emphasis}, `MW.' = Missing Words = words from the evaluation set that could not be formed, `SE' = start/end tokens, `Diff.' = the difference between the scores with and without (\tabref{tb:en_emb_results}) the syllable splitter model.}
\label{tb:en_eval_with_syllsplit_results}
\ra{1.2}
\begin{tabular}{@{}rcccccccccccccc@{}}
\toprule
&& MW. & Diff. & WS353 & Diff. & WS353/3ct & Diff. && MW. & Diff. & SE2017 & Diff. & SE2017/3ct & Diff. \\
\midrule
\emph{Attention 1} &&&& &&& &&&& &&& \\
SE: No   && 1 & $-$1 & 0.514 & $+$.001 & 0.343 & $+$.0   && 31 & $-$44 & 0.320 & $-$.117 & 0.339 & $-$.004 \\
SE: 10\% && 1 & $-$1 & 0.606 & $+$.002 & 0.364 & $-$.003 && 32 & $-$43 & 0.371 & $-$.129 & 0.425 & $+$.040 \\
SE: 25\% && 1 & $-$1 & 0.634 & $+$.002 & 0.399 & $+$.003 && 35 & $-$40 & 0.382 & $-$.136 & 0.419 & $+$.009 \\
SE: 50\% && 1 & $-$1 & 0.624 & $+$.000 & 0.396 & $-$.004 && 36 & $-$39 & 0.416 & $-$.100 & 0.437 & $-$.009 \\
SE: 75\% && 1 & $-$1 & 0.631 & $+$.000 & 0.396 & $-$.003 && 36 & $-$39 & 0.395 & $-$.133 & 0.433 & $+$.006 \\
SE: 100\% && 1 & $-$1 & 0.629 & $+$.000 & 0.406 & $+$.002 && 36 & $-$39 & 0.409 & $-$.029 & 0.432 & $-$.005 \\
\midrule
\emph{Attention 2} &&&& &&& &&&& &&& \\
SE: No   && 1 & $-$1 & 0.500 & $+$.002 & 0.341 & $-$.001 && 31 & $-$44 & 0.314 & $-$.142 & 0.323 & $-$.008 \\
SE: 10\% && 1 & $-$1 & 0.584 & $+$.004 & 0.364 & $+$.001 && 32 & $-$43 & 0.379 & $-$.109 & 0.398 & $+$.004 \\
SE: 25\% && 1 & $-$1 & 0.605 & $+$.005 & 0.356 & $-$.002 && 35 & $-$40 & 0.347 & $-$.160 & 0.418 & $+$.043 \\
SE: 50\% && 1 & $-$1 & 0.566 & $+$.002 & 0.381 & $+$.004 && 36 & $-$39 & 0.382 & $-$.137 & 0.419 & $+$.000 \\
SE: 75\% && 1 & $-$1 & 0.592 & $+$.004 & 0.390 & $+$.002 && 36 & $-$39 & 0.398 & $-$.119 & 0.408 & $-$.011 \\
SE: 100\% && 1 & $-$1 & 0.602 & $+$.004 & 0.383 & $+$.003 && 36 & $-$39 & 0.394 & $-$.109 & 0.399 & $-$.021 \\
\bottomrule
\end{tabular}
\end{table}
\end{landscape}

\begin{landscape}
\begin{table}[!htbp]
\centering
\caption{Accuracy results for different embedding methods: Dutch. Model names in \emph{emphasis}, `MW.' = Missing Words = words from the evaluation set that could not be formed, `SE' = start/end tokens, `Diff.' = the difference between the scores with and without (\tabref{tb:nl_emb_results}) the syllable splitter model.}
\label{tb:nl_eval_with_syllsplit_results}
\ra{1.2}
\begin{tabular}{@{}rcccccccccccccc@{}}
\toprule
&& MW. & Diff. & WS353 & Diff. & WS353/3ct & Diff. && MW. & Diff. & SE2017 & Diff. & SE2017/3ct & Diff. \\
\midrule
\emph{Attention 1} &&&& &&& &&&& &&&  \\
SE: No   && 7 & $-$13 & 0.436 & $+$.018 & 0.381 & $+$.014 && 73 & $-$101 & 0.429 & $+$.035 & 0.408 & $+$.033 \\
SE: 10\% && 7 & $-$13 & 0.460 & $+$.018 & 0.404 & $+$.018 && 73 & $-$101 & 0.452 & $+$.037 & 0.408 & $+$.037 \\
SE: 25\% && 7 & $-$13 & 0.480 & $+$.017 & 0.413 & $+$.011 && 73 & $-$101 & 0.456 & $+$.039 & 0.409 & $+$.028 \\
SE: 50\% && 7 & $-$13 & 0.484 & $+$.014 & 0.396 & $+$.020 && 73 & $-$101 & 0.478 & $+$.032 & 0.427 & $+$.039 \\
SE: 75\% && 7 & $-$13 & 0.492 & $+$.020 & 0.351 & $+$.013 && 75 & $-$99 & 0.462 & $+$.022 & 0.423 & $+$.025 \\
SE: 100\% && 7 & $-$13 & 0.480 & $+$.021 & 0.393 & $+$.016 && 79 & $-$95 & 0.453 & $+$.031 & 0.420 & $+$.019 \\
\midrule
\emph{Attention 2} &&&& &&& &&&& &&&  \\
SE: No   && 7 & $-$13 & 0.443 & $+$.025 & 0.411 & $+$.023 && 73 & $-$101 & 0.410 & $+$.023 & 0.401 & $+$.021 \\
SE: 10\% && 7 & $-$13 & 0.439 & $+$.019 & 0.373 & $+$.020 && 73 & $-$101 & 0.407 & $+$.039 & 0.395 & $+$.033 \\
SE: 25\% && 7 & $-$13 & 0.482 & $+$.021 & 0.394 & $+$.020 && 73 & $-$101 & 0.377 & $+$.012 & 0.420 & $+$.017 \\
SE: 50\% && 7 & $-$13 & 0.464 & $+$.014 & 0.368 & $+$.020 && 73 & $-$101 & 0.388 & $+$.031 & 0.429 & $+$.034 \\
SE: 75\% && 7 & $-$13 & 0.493 & $+$.020 & 0.380 & $+$.016 && 75 & $-$99 & 0.376 & $+$.025 & 0.409 & $+$.005 \\
SE: 100\% && 7 & $-$13 & 0.479 & $+$.018 & 0.369 & $+$.012 && 79 & $-$95 & 0.389 & $+$.021 & 0.416 & $+$.024 \\
\bottomrule
\end{tabular}
\end{table}
\end{landscape}

The network architecture we used was a sequence-to-sequence Transformer network \cite{transformer}, as implemented by PyTorch. We experimented with several configurations, the parameters and results of which are shown in \tabref{tb:syll_split_results}. We chose to constrain the number of heads to be at most one fourth of the embedding dimension. One thing we could not get properly working however, was training with batches. We found that the model would not converge whenever batches with size $>$1 were used. Hence, all models were trained using only one sample at a time. On disk model sizes varied from 65kB for the smallest one to 1MB for the largest one.

From these results, we gather that the parameter with the biggest impact on performance is the embedding size, while the impact of the other parameters (number of layers, number of heads and hidden size) is not always consistent. As evidenced by the results, the performance of the Dutch models is markedly better than the English models. No doubt this is partly due to the Dutch training set being considerably larger, although we also suspect differences in splitting rules between both languages might have an effect. Nevertheless, at 78.6\% and 90.3\% accuracy for English and Dutch respectively, it is clear that these models perform quite to very well.

We then proceeded to training a model for each language using all available data. For this, we settled on the following parameters (for both models): $layers=1$, $embedding=64$, $hidden=256$ and $heads=8$. These models were then used when evaluating our embeddings as follows:
\begin{enumerate}
\item When the syllabic decomposition of a word was unknown (i.e., not contained in either our Dutch Wiktionary dataset or English scraped dataset), we would use the splitter model to obtain a decomposition.
\item We would then check that the predicted decomposition did not contain any spurious characters. (Since the splitter model is a seq-2-seq model, it can and does happen that characters are either removed or added when they should not be.) In other words, when joining back together all syllables we should obtain the original word. If not, we would ignore the decomposition, and skip the word.
\item Lastly, we would check that the syllabic decomposition contained only syllables for which we have an embedding. If not, we would skip the word.
\item If on the other hand all syllables were known, we would use the decomposition to create a word embedding the same way we would if we had obtained the decomposition from our dataset of known decompositions.
\end{enumerate}

We only performed this experiment for the ``regular" and ``/3ct" Attention 1 and 2 models. Tables \ref{tb:en_eval_with_syllsplit_results} and \ref{tb:nl_eval_with_syllsplit_results} show the results, as well as the difference with the same runs without using the splitter model (and thus only using a table lookup).

These results unfortunately cement our earlier observation that our embeddings perform rather poorly on OOV problems. Despite being able to resolve significantly more words by using our splitter models, the evaluation scores are not that much affected. Worse, for the English language performance is mostly adversely affected. Results for Dutch often do increase with about 4\%, which is somewhat encouraging. We reiterate our belief that this poor perfomance is strongly related to data sparseness and the fact that we only use one possible decomposition to create an embedding vector.

\section{Conclusion}
\label{s:conclusion}
We translated two word embedding evaluation sets, WordSim353 and SemEval 2017, from English to Dutch, and researched extensively the possibility of using syllables in both languages to compress larger word embedding models -- ConceptNet Numberbatch in our specific case. Our goal was two-fold: to obtain a much smaller model that would still perform adequately, and to provide an alternative to the common $n$-gram solution to the OOV problem.

The results are somewhat of a curate's egg. On the one hand, averaged over the English and Dutch cases, we managed to reduce the model size by a factor of 20 while retaining 80\% of the original model performance on the Wordsim353 and SemEval 2017 evaluation sets. This is clearly an encouring result. On the other hand, we found there to be significant issues with data sparseness leading to poor OOV performance. Moreover, our syllable-based models are outperformed by existing $n$-gram models. On the flip side, training models is considerably faster, taking only minutes instead of hours.

Nevertheless, we do feel investigating the use of syllable embeddings presents an interesting avenue, because of the inherent elegance of the idea. Syllables are after all, after letters, the natural ``building blocks" of words. In order to further improve upon this work, we strongly believe the most promising route is to investigate the usage of more than a single syllabic decomposition to compose the word embeddings. In our work, we only looked at the ``true" syllabic decomposition of a word. By combining different syllables in multiple ways to obtain a same word, one could get closer to the way $n$-grams work. E.g., the true syllabic decomposition of ``apple" is ``ap-ple", but with the syllables that are known from the data we used could also be decomposed as ``app-le", ``app-l-e", etc. However, as we point out in the main text, the issue with this approach is that the number of possible decompositions grows exponentially with increasing wordlength, with the Dutch word ``aansprakelijkheidsverzekeringen" yielding no less than 3,179,520 possible decompositions. The issue then becomes: how many possibilities should be kept at most, wich ones, and why exactly those and not others?


\begin{acknowledgements}
This work has been made possible by the Flanders Innovation \& Entrepreneurship TETRA project ``Start To Deep Learn". We would further like to thank Radix (\url{https://radix.ai/}) for providing us with the research question and technical support.
\end{acknowledgements}


\bibliographystyle{splncs04}

\end{document}